%% file: main.tex
\documentclass[10pt,twocolumn,letterpaper]{article}

\usepackage{iccv}              

\definecolor{iccvblue}{rgb}{0.21,0.49,0.74}
\usepackage[pagebackref,breaklinks,colorlinks,allcolors=iccvblue]{hyperref}
\usepackage{pifont}
\usepackage[utf8]{inputenc}
\input{0_config/00_packages}
\input{0_config/00_macros}
\input{0_config/00_const}

\title{\TITLE} %

\input{0_config/00_authors.tex}

\begin{document}

\twocolumn[{%
\renewcommand\twocolumn[1][]{#1}%
\maketitle
\input{2_tex_FIG/teaser}

}
]

\input{main_paper}

{
    \small
    \bibliographystyle{ieee_fullname}
    \bibliography{main}
}

\end{document}

%% file: 0_config/00_packages.tex
\usepackage{graphicx}
\usepackage{float}

\usepackage{lipsum}  
\usepackage{bbm}

%% file: 0_config/00_macros.tex
\definecolor{mypurple}{RGB}{0, 128, 255}
\definecolor{dexcolor}{RGB}{0, 117, 4}
\definecolor{honotatecolor}{RGB}{1, 0, 253}

\newcommand{\colorRef}[1]{\textcolor{black}{#1}} %

\newcommand{\refFig}[1]{\mbox{\colorRef{Figure~\ref{#1}}}}

\newcommand{\refTab}[1]{\mbox{\colorRef{Table~\ref{#1}}}}

\newcommand{\refEq}[1]{\mbox{\colorRef{Equation~\ref{#1}}}}
\newcommand{\refsec}[1]{\colorRef{Sec.~\ref{#1}}}

\newcommand{\subtitle}[1]{\textbf{#1}.}

\definecolor{GreenColor}{rgb}{0.137,0.573,0.565}
\definecolor{OrangeColor}{rgb}{0.914,0.541,0.0.141}
\definecolor{PurpleColor}{rgb}{0.5,0,0.7}
\definecolor{BlueColor}{rgb}{0,0.725,0.949}
\definecolor{PinkColor}{rgb}{0.9843,0.19215,0.6}

\newcommand{\V}[1]{\mathbf{#1}} %
\newcommand{\R}{\rm I\!R}

\newcommand{\myparagraph}[1]{\noindent\textbf{#1:}}

\newcommand\blfootnote[1]{%
  \begingroup
  \renewcommand\thefootnote{}\footnote{#1}%
  \addtocounter{footnote}{-1}%
  \endgroup
}

%% file: 0_config/00_const.tex
\newcommand{\nameCOLOR}[1]{\textcolor{black}{#1}} %
\newcommand{\methodname}{\mbox{\nameCOLOR{MagicHOI}}\xspace}

\newcommand{\nvsm}{\mbox{\nameCOLOR{Zero-1-to-3}}\xspace}
\newcommand{\nvs}{NVS\xspace}

\newcommand{\TITLE}{
  MagicHOI: Leveraging 3D Priors for Accurate Hand-object Reconstruction from Short Monocular Video Clips%
}

\newcommand{\suppl}{\textcolor{black}{SupMat}\xspace}

\newcommand{\rgb}{RGB\xspace}

\newcommand{\twoD}{{2D}\xspace}

\newcommand{\threeD}{\xspace{3D}\xspace}

\newcommand{\sota}{{SOTA}\xspace}

\newcommand{\groundtruth}{{ground-truth}\xspace}

%% file: 0_config/00_authors.tex
\author{
Shibo Wang$^{1}$ 
\quad Haonan He$^{1}$ 
\quad Maria Parelli$^{3,\dagger}$
\quad Christoph Gebhardt$^{1}$
\quad Zicong Fan$^{3,4}$ 
\quad Jie Song$^{1,2}$ \\
 $^1$The Hong Kong University of Science and Technology (Guangzhou) \\
 $^2$The Hong Kong University of Science and Technology \\
 $^3$ETH Z{\"u}rich, Switzerland
 \quad $^4$Max Planck Institute for Intelligent Systems, T{\"u}bingen, Germany \\
}

%% file: 2_tex_FIG/teaser.tex
\begin{center}
    \centerline{\includegraphics[width=0.9\linewidth]{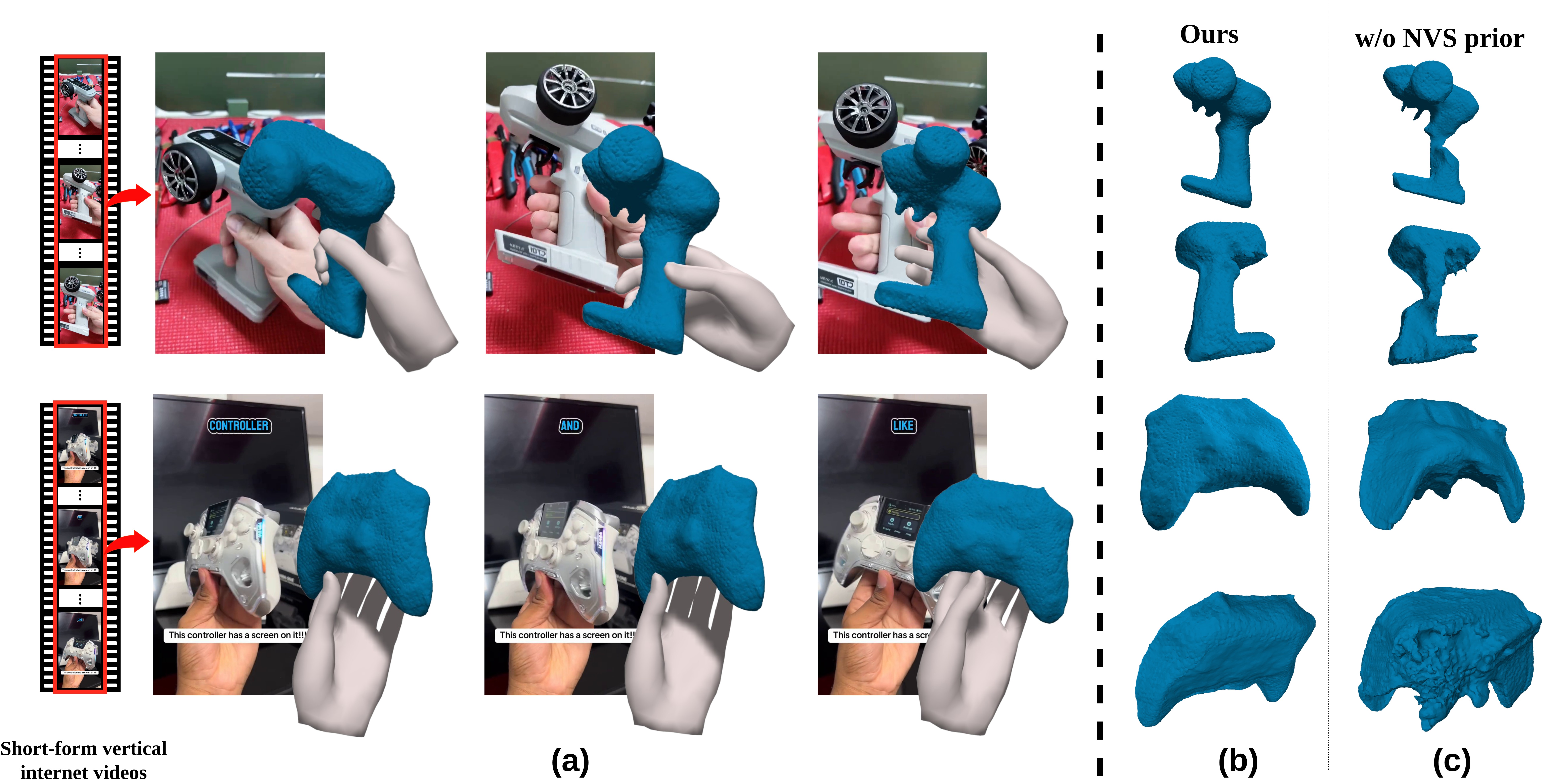}}
    \captionof{figure}{\textbf{\methodname:} Given a short-form monocular video capturing hand-object interaction, our method reconstructs high-quality \threeD hand and object surfaces, including occluded object regions caused by hand interaction or object self-occlusion. 
    \textbf{(a)} Input images and corresponding reconstructed surfaces of the hand and object. 
    \textbf{(b, c)} Comparison of our method with and without a novel view synthesis (\nvs) prior from the object's front and back views, demonstrating improved reconstruction in occluded areas.
    Best viewed in color.
    }
    \label{fig:teaser}
\end{center}%

%% file: main_paper.tex
\input{4_tex_Paper/0_abstract}

\section{Introduction}
\label{sec:intro}

\input{4_tex_Paper/1_intro}

\section{Related Work}
\label{sec:related_work}
\input{4_tex_Paper/2_related_work}

\section{Method: \methodname}
\label{sec:method}
\input{4_tex_Paper/3_method}

\section{Experiments}
\label{sec:exp}

\input{4_tex_Paper/4_experiments}

\section{Conclusion}
\label{sec:conclusion}
\input{4_tex_Paper/5_conclusion}

%% file: 4_tex_Paper/0_abstract.tex
\blfootnote{$\dagger$ Prior to joining University of T\"ubingen and T\"ubingen AI Center }

\begin{abstract}

Most RGB-based hand-object reconstruction methods rely on object templates, while template-free methods typically assume full object visibility. This assumption often breaks in real-world settings, where fixed camera viewpoints and static grips leave parts of the object unobserved, resulting in implausible reconstructions.
To overcome this, we present \methodname, a method for reconstructing hands and objects from short monocular interaction videos, even under limited viewpoint variation. Our key insight is that, despite the scarcity of paired 3D hand-object data, large-scale novel view synthesis diffusion models offer rich object supervision. This supervision serves as a prior to regularize unseen object regions during hand interactions.
Leveraging this insight, we integrate a novel view synthesis model into our hand-object reconstruction framework. 
We further align hand to object by incorporating visible contact constraints.
Our results demonstrate that \methodname significantly outperforms existing state-of-the-art hand-object reconstruction methods. We also show that novel view synthesis diffusion priors effectively regularize unseen object regions, enhancing 3D hand-object reconstruction. 
\end{abstract}

%% file: 4_tex_Paper/1_intro.tex
Humans interact effortlessly with a wide variety of objects every day, often without conscious thought. These complex interactions are frequently captured in casual videos, which are abundant on the internet (e.g., product reviews on TikTok, Instagram Reels, and YouTube Shorts). Such videos provide a valuable resource for scaling human demonstrations in robotic grasping~\cite{christen2022dgrasp,zhang2024artigrasp,zhang2024graspxl}.
A key challenge in reconstructing 3D hand-object interactions at scale is that most internet videos do not offer complete, dense views of the objects due to hand-induced occlusions and object self-occlusions, particularly when the objects are not fully rotated. This incomplete observation hinders the accurate reconstruction of hands and objects (see \refFig{fig:teaser}).

To address this challenge, we introduce the task of \textit{Partially-visible Hand-object Reconstruction}, which aims to reconstruct 3D hand and object geometries from videos where the objects are not fully visible from all angles. Specifically, we focus on short video clips of interactions, as these often contain under-observed regions of the objects (e.g., the action of picking up a cup of coffee captured from a static monocular view).

Existing methods for hand-object reconstruction typically rely on pre-scanned 3D object templates~\cite{tekin2019ho, corona2020ganhand, liu2021semi, cao2021handobject, yang2021cpf,fan2024benchmarks}, which limit their scalability in real-world scenarios. While emerging learning-based approaches are available, most depend on supervised training using a limited number of paired 3D hand-object annotations, resulting in poor generalization~\cite{hasson2019obman, karunratanakul2020graspField, ye2022s}. Recently, methods leveraging volumetric rendering have shown promise in generalizing to unseen objects~\cite{fan2024hold, huang2022reconstructing, hampali2023inhand}. However, these approaches rely heavily on RGB supervision and struggle to accurately reconstruct hands and objects when parts of the object are occluded or only partially visible (see \refFig{fig:sota_compare}).

In this paper, we introduce \methodname, a method designed to address restricted viewpoints and significant occlusions in realistic hand-object interaction scenarios. \methodname takes as input a monocular video and reconstructs the 3D surfaces of both the hand and the object, even in short video sequences with partially visible object regions. Our key insight is that, although 3D hand-object annotations are limited, foundational models such as novel view synthesis (\nvs) diffusion models provide abundant object supervision~\cite{liu2023zero1to3}. This external supervision can be utilized as a prior to help regularize unseen or occluded object regions during hand-object interactions.

In detail, \methodname first employs structure-from-motion (SfM) to obtain the initial camera poses of the object from the input video. Once these observed views are calibrated through SfM, we utilize volumetric rendering to reconstruct the object geometry using an implicit signed distance field (SDF)~\cite{yariv2021volume,muller2022instant}.
To regularize the reconstruction in unobserved regions, we sample novel views and apply a score-distillation sampling loss using an \nvs diffusion model~\cite{liu2023zero1to3}. However, randomly sampling novel views to enforce a prior may negatively impact the reconstruction quality of object regions that are visible.
To address this, we introduce a visibility-aware weighting strategy. Finally, \methodname aligns the hand and the object to the same space while incorporating hand-object interaction constraints.

We qualitatively and quantitatively demonstrate that our visibility-aware weighting strategy, combined with the integration of an \nvs diffusion prior, significantly enhance 
the hand-object reconstruction quality in the partially-visible hand-object reconstruction setting, surpassing the performance of the state-of-the-art methods. To highlight the robustness of \methodname in challenging real-world scenarios, we also illustrate its ability to faithfully reconstruct short-form videos sourced from online content.

In summary, our contributions are as follows: 1) We present \methodname, a holistic framework that effectively leverages an \nvs diffusion prior for accurate 3D hand-object reconstruction from short monocular video sequences, even in the presence of unobserved parts of the hand and object; 2) We demonstrate that the \nvs model can be utilized to regularize template-free hand-object reconstruction; 3) We introduce a novel visibility-aware weighted sampling strategy to balance the regularization between the observed and unobserved object regions using this prior; 4) We evaluate our method both qualitatively and quantitatively, demonstrating its superior performance compared to the state-of-the-art methods and achieving realistic reconstruction results.
Our project website can be found here: 
 \href{https://byran-wang.github.io/MagicHOI}{\texttt{byran-wang.github.io/MagicHOI}}.

%% file: 4_tex_Paper/2_related_work.tex
\myparagraph{Hand pose and shape recovery}
The task of reconstructing hands in \threeD is a long-standing problem in computer vision, where most methods focus on single-hand reconstruction~\cite{iqbal2018hand,Mueller2018ganerated,Spurr2018crossmodal,spurr2021peclr,spurr2020eccv,zimmermann2017iccv,Boukhayma2019,ziani2022tempclr,fan2021digit,simon2017hand,Zhang2019endtoend,interhand,li2022interacting,zhang2021interacting,tzionas2013directional,chen2023handavatar,fu2023deformer}. 
In particular, Zimmermann \etal~\cite{zimmermann2017iccv} introduce the first convolutional networks for hand pose estimation.
Spurr \etal~\cite{spurr2020eccv} present bio-mechanical constraints allowing for semi-supervised learning on in-the-wild images. 
Contrastive learning objectives are also explored for learning hands in action on images without labels~\cite{spurr2021peclr,ziani2022tempclr}.
With the introduction of the InterHand2.6M dataset by Moon \etal~\cite{interhand}, the interacting hand pose estimation research has exploded~\cite{moon2023bringing,lee2023im2hands,li2022interacting,meng20223d,interhand,moon2023dataset,li2023renderih,ohkawa2023assemblyhands,guo2023handnerf,tse2023spectral,fan2021digit}. 
For example, Tse \etal~\cite{tse2023spectral} introduced a spectral graph transformer network for estimating the surfaces of two strongly interacting hands. 
Pavlakos \etal~\cite{pavlakos2024reconstructing} train a vision transformer to achieve robust monocular hand pose estimation under unconstrained conditions.
Yu \etal~\cite{yu2025dynhamr} propose Dyn-HaMR, which reconstructs interacting hand poses in a global frame by combining SLAM with a learned hand motion prior~\cite{duran2024hmp} for spatial alignment.
In contrast to these approaches, we focus on joint hand-object \threeD reconstruction.

\myparagraph{Template-based hand-object reconstruction}
Reconstructing hand-object interaction is challenging and has gained increasing attention in recent years~\cite{hasson2019obman,liu2021semi,yang2021cpf,grady2021contactopt,Hasson2020photometric,tekin2019ho,corona2020ganhand,zhou2020monocular,hasson2021towards,tse2022collaborative,yang2021cpf,liu2024egohdm}.
Most existing methods assume a known object template and estimate only hand and object poses~\cite{tekin2019ho,corona2020ganhand,liu2021semi,cao2021handobject,yang2021cpf,fan2024benchmarks}.
For example, 
Yang \etal~\cite{yang2021cpf} introduce a contact potential field to improve hand-object contact.
Liu \etal~\cite{liu2021semi} propose a transformer-based contextual reasoning module that captures the synergy between hand and object features, yielding stronger responses at contact regions.
Zhou \etal\cite{zhou2022toch} learn an interaction motion prior to refine noisy motion estimates from a single-frame hand-object reconstruction approach.
Fan \etal~\cite{fan2023arctic,fan2024benchmarks} introduce the first approach to jointly reconstruct two dexterous hands and articulated objects from \rgb inputs.
Despite accurate pose estimation, these methods struggle with novel objects and in-the-wild videos due to dependence on known templates.

\myparagraph{Template-free hand-object reconstruction}
Recent advancements have introduced more generalizable techniques~\cite{swamy2023showme, Chen2022TensoRF, qu2023novel, huang2022reconstructing, fan2024hold, hasson2019obman, ye2022hand, chen2023gsdf, ye2022s, ye2023diffhoi, ye2024ghop, wu2024mccho}, incorporating differentiable rendering, data-driven priors, and compositional implicit representations. However, these methods often impose constraints such as requiring rigid hand-object interactions~\cite{huang2022reconstructing, Prakash2023ARXIV}, multi-view observations~\cite{qu2023novel}, category-level hand-object supervision~\cite{hasson2019obman, ye2022hand, chen2023gsdf, ye2022s, ye2023diffhoi, ye2024ghop, wu2024mccho}, or complete object observations~\cite{fan2024hold}. Consequently, they struggle with occlusions, leading to incomplete object surfaces in cases of hand-induced or self-occlusions.
In contrast, our approach integrates 3D priors with geometry information using a visibility-aware weighting strategy, enhancing object reconstruction quality in both occluded and observed regions, even in extremely short video clips.

\myparagraph{\threeD object reconstruction}
Reconstructing objects from images has been a long-standing challenge in computer vision. Structure from Motion (SfM) ~\cite{hartley2003multiple, scharstein2002taxonomy, szeliski2022computer, tomasi1992shape} has remained a robust solution, particularly for multiview reconstruction and camera parameter estimation. However, traditional SfM methods are limited to scenarios where the object is either fully visible or nearly unoccluded in the input views.
To address the reconstruction of occluded object shapes, many research efforts have turned to learning-based approaches. These approaches leverage large-scale 3D object datasets, such as Objaverse ~\cite{deitke2023objaverse} and Objaverse-XL ~\cite{deitke2023objaversexl}, and include methods like regression ~\cite{li2020regression}, retrieval ~\cite{tatarchenko2019retrieval}, diffusion models ~\cite{liang2024luciddreamer, liu2023zero1to3, poole2022dreamfusion, liu2023syncdreamer, long2024wonder3d, shi2023mvdream}, and large-scale reconstruction models ~\cite{hong2023lrm, tang2024lgm, wang2023pf, wang2024crm}.
Despite progress, standard object reconstruction methods underperform with hand-object interaction images due to hand occlusions. Unlike these, \methodname jointly reconstructs hands and objects from videos.

%% file: 4_tex_Paper/3_method.tex
\refFig{fig:method} provides an overview of our approach, \methodname, which reconstructs hand--object interactions from a \rgb video captured with limited viewpoints.
We first initialize the hand and object poses for every frame (\refsec{sec:pose_init}).
Next, we integrate a novel view synthesis (\nvs) model into the pipeline and align its coordinate frame with that of the object (\refsec{sec:integrating_vsm}).
With the \nvs model in place, we optimize an implicit signed distance field (SDF) that regularizes the reconstruction of object regions that are not directly observed (\refsec{sec:joint_optimization}).
Finally, we align the hand to the object using visible hand-object contact constraints (\refsec{sec:hand_ref}).

\subsection{Initialization}
\label{sec:pose_init}
\myparagraph{Hand initialization} For each image ${I}^\text{obs} \in \mathbb{R}^{3{\times}H{\times}W}$ from a short video clip, we leverage an off-the-shelf hand pose estimator~\cite{pavlakos2024HaMeR} to obtain MANO~\cite{mano} parameters, which include the hand pose $\theta \in \mathbb{R}^{45}$, shape $\beta \in \mathbb{R}^{10}$, global rotation $\mathbf{R}_h \in \mathrm{SO}(3)$, and translation $\mathbf{t}_h \in \mathbb{R}^3$. 

\myparagraph{Object initialization} Following~\cite{fan2024hold}, we apply HLoc~\cite{sarlin2019coarse, sarlin2020superglue} to perform structure-from-motion (SfM), obtaining the camera intrinsics ${\V{K}} \in \mathbb{R}^{3 \times 3}$, rotation $\mathbf{R}^{\text{SfM}} \in \mathrm{SO}(3)$, and translation $\mathbf{t}^{\text{SfM}} \in \mathbb{R}^3$. For each frame, we additionally generate a depth map ${\V{D}} \in \mathbb{R}^{H \times W}$ using multi-view stereo (MVS)~\cite{he2023dfsfm, schoenberger2016mvs}. These steps are applied to images segmented by the off-the-shelf model Cutie~\cite{cheng2023cutie}. In fact, advances in learning-based pose estimation, such as MASt3R~\cite{leroy2024mast3r}, can provide more robust object pose estimates without relying on SfM convergence.

\subsection{Novel view synthesis model}
\label{sec:integrating_vsm}
A key contribution of \methodname is leveraging an \nvs model \nvsm to reconstruct objects with hands even in occluded object regions.

\myparagraph{Inpainting}
Before using the \nvs model to supervise our reconstruction, one needs to condition it on a reference image ${I}^\text{ref}$  where only the object is visible.
To achieve this, we extract an object-only image based on the object segmentation mask and perform inpainting on the hand region to create the object-only reference image ${I}^\text{ref}$.
We use InpaintAnything~\cite{yu2023inpaint} to inpaint the hand region.
The inpainted image $\tilde{I}^\text{ref}$ is then centered, cropped, and resized to suit the input format of the \nvs model ${I}^\text{ref}$.
We select the reference frame from the input video based on a simple heuristic. Specifically, we use off-the-shelf segmentation to extract hand and object masks for each frame, compute the object-to-hand pixel ratio, and choose the frame with the highest ratio. This automated selection method prioritizes views where the object is more clearly visible.

\myparagraph{Novel view synthesis model}
We use Zero-1-to-3~\cite{liu2023zero1to3} as our view synthesis backbone. 
In a nutshell, conditioned on an input image ${I}^\text{ref}$, its camera pose $\V{y}^\text{ref}$ and a target camera view $\V{y}$,
the \nvs model $f_\text{\nvs}({I}^\text{ref}, \V{y}^\text{ref}, \V{y})$ generates an image $I^\text{\nvs}$ for the novel view  $\V{y}$ where the view $\V{y} = [\varphi, \phi, \rho]$ represents the camera pose in spherical coordinates, $\varphi$ denotes elevation, $\phi$ denotes azimuth, and $\rho$ denotes distance from the object center. 
The camera intrinsic of the reference image ${I}^\text{ref}$ is represented by ${\V{K}^\text{ref}} \in \mathbb{R}^{3 \times 3}$, centered on the image plane and characterized by a predefined focal length. See more details in \suppl.

We use the pre-trained \nvsm model with frozen weights in our framework. It is trained using a forward diffusion process that progressively corrupts a novel-view image with Gaussian noise, and a denoiser $\epsilon_\omega$ is then learned to reconstruct the image by reversing this process~\cite{ho2020denoising}. Training is performed on large-scale synthetic data rendered from 3D object models. For full architectural and training details, we refer to the original \nvsm paper~\cite{liu2023zero1to3}.

\myparagraph{Space alignment}
To leverage both observations from \rgb video input and object geometry prior in \nvsm, we must align our SfM object space to that of the \nvsm object space. 
We use the object space from \nvsm as our canonical object space and align the SfM object space to it.
To achieve this, we first extract the 2D correspondences $\{\V{p}^{\text{ref}},\tilde{\V{p}}^{\text{ref}}\}$ 
between the reference image ${I}^\text{ref}$ and the same image but in the original resolution $\tilde{I}^\text{ref}$.
Since the \nvsm model can render the depth map \(\V{D}^{\text{ref}}\) of the reference view and the MVS process provides a depth map \(\tilde{\V{D}}^{\text{ref}}\),
we apply inverse perspective projection to obtain \threeD correspondences $\{\V{P}^\text{ref}, \tilde{\V{P}}^\text{ref}\}$ using the intrinsics from the \nvsm model and the SfM estimates respectively.

Finally, we estimate the rigid transformation that brings the
SfM object frame into the \nvsm frame by solving the rotation
$\V{R}_{\text{a}}\!\in\!\mathrm{SO}(3)$ and translation
$\V{t}_{\text{a}}\!\in\!\mathbb{R}^3$. The optimization objective combines a
3‑D correspondence term $\mathcal{L}_{3D}$ and a Perspective‑\(n\)-Point (PnP) term $\mathcal{L}_{2D}$:
\begin{align}
\mathcal{L}_\text{a} &=
     \mathcal{L}_\text{3D} + \lambda_\text{2D}\mathcal{L}_\text{2D},
\end{align}
where $\lambda_\text{2D}=1.0$. See more details in \suppl.

Using the estimated rigid transformation \((\mathbf{R}_{\text{a}},\mathbf{t}_{\text{a}})\),  
we convert each SfM camera pose \((\mathbf{R}^{\text{SfM}},\mathbf{t}^{\text{SfM}})\)  
from the SfM frame to the \nvsm frame:
\begin{align}
    \mathbf{R} = \mathbf{R}_\text{a} \mathbf{R}^{\text{SfM}},
    \quad
    \mathbf{t} = \mathbf{R}_\text{a} \mathbf{t}^{\text{SfM}} + \mathbf{t}_\text{a}.
\end{align}
where \(\mathbf{R}\in\mathrm{SO}(3)\) and
\(\mathbf{t}\in\mathbb{R}^{3}\).

\input{2_tex_FIG/method}

\subsection{Occlusion-robust object reconstruction}
\label{sec:joint_optimization}
With the help of the \nvs backbone, we can synthesize novel views to reveal parts of the object that are occluded in the original observations, thereby enabling occlusion-robust object reconstruction.

We represent the object using an implicit neural signed distance and texture field $f_{\psi_o}$, which can be volumetrically rendered into RGB images and defined as:
\begin{align}
    f_{\psi_o}(\mathbf{x}): \mathbb{R}^{3} &\rightarrow \mathbb{R} \times \mathbb{R}^{3}
\end{align}
where $f_{\psi_o}$ maps a spatial point $\mathbf{x}$ to its signed distance to the object surface and its corresponding texture color. The function is parameterized by learnable parameters $\psi_o$.

\myparagraph{Supervision on the observed object regions}
For observed \rgb images $\{I^\text{obs}\}$, we supervise our object model $\psi_o$ with image evidence, namely \rgb images and segmentation masks $\mathcal{L}_\text{RGB} + \lambda_\text{segm}\mathcal{L}_\text{segm}$
where $\mathcal{L}_\text{RGB}$ enforces photometric consistency between the rendered and observed object images, $\mathcal{L}_\text{segm}$ encourages alignment with the object segmentation mask (see more details in \suppl).

\myparagraph{Supervision on the unobserved object regions}
To supervise unobserved object regions using the object geometry prior, we adopt the Score Distillation Sampling (SDS) loss~\cite{poole2022dreamfusion}. 
A novel view image $I^\text{\nvs}$ is rendered from the object model $f_{\psi_o}$ using a randomly sampled camera pose $\V{y}$. 
We sample a diffusion timestep $t \sim \mathrm{Uniform}(\{0, \ldots, T\})$ and obtain the corresponding Gaussian noise $\epsilon_t$. 
A noisy image $\V{z}_t$ is computed by adding the noise $\epsilon_t$ to the rendered image  $I^\text{\nvs}$. 
Given the known noise $\epsilon_t$ from the step $t$ of the forward diffusion process, we supervise the model by minimizing the discrepancy between it and the \nvsm denoiser's prediction:
\begin{align}
\mathcal{L}_\text{NVS}=  w(t) \left\lVert \epsilon_t - \epsilon_\omega(\V{z}_t; t, \mathcal{C}) \right\rVert^2
\end{align}
where $\epsilon_\omega$ is the denoising network of the \nvsm model, $\mathcal{C} = \Psi({I}^\text{ref}, \V{y}^\text{ref}, \V{y})$ is the embedding of the condition. 
The scalar $w(t)\in \R$ is a weighting function predefined by the diffusion model and the step $t$.

\myparagraph{Visibility-aware weighting strategy}
\label{sec:weighting_strategy}
To balance image evidence and object geometry prior from \nvsm, we devise a visibility-aware weighting strategy.
In particular, after the object model has been optimized using observed views for some iterations, we derive a coarse 3D visibility grid $O(x, y, z)$ from the underlying \threeD object representation, which is a coarse voxel representation around the object.
For each pixel of the observed views, we cast a ray from an object pixel into the visibility grid to determine the occluded voxels for novel views.
This process then labels a voxel as visible (valued 1) if the voxel is the first intersecting point of the ray. 
If a voxel is never the first intersected point of all rays, it is labeled as non-visible (valued 0).


For each novel view $\V{y}$, we rasterize the visibility grid $O$ to produce a visibility image $V(u, v)$, where each pixel takes a value from $\{1, 0, -1\}$: $1$ indicates the voxel is observed by at least one ray in the observed input images, $0$ indicates it has never been observed, and $-1$ denotes background.
From $V(u, v)$, we compute the visibility ratio $\beta$ for the given view $\V{y}$.
In detail, we compute the ratio of pixels in observed regions to those in unobserved regions
\begin{align}
\beta = \frac{\sum\limits_{(u, v) \in V} \mathbb{I}[V(u, v) = 1]}{\sum\limits_{(u, v) \in V} \mathbb{I}[V(u, v) = 0]}
\end{align}
Given the ratio $\beta$, we compute a weighting factor as $\mu = e^{-\beta^2/0.6}$ for a randomly sampled novel view $\V{y}$.

\myparagraph{Total loss}
To summarize, at each iteration, we sample one \rgb image from the video input images $\{I^\text{obs}\}$, the reference image $I^\text{ref}$, and multiple novel views $\{\V{y}\}$, and then compute the total loss $\mathcal{L}_o$ for learning the object network parameters $\psi_o$
where

\noindent\resizebox{\linewidth}{!}{%
  \parbox{\linewidth}{%
    \begin{align}
        \mathcal{L}_o = \mathcal{L}_\text{RGB} + \lambda_\text{segm} \mathcal{L}_\text{segm} + \lambda_\text{smooth} \mathcal{L}_\text{smooth} + \mu \mathcal{L}_\text{\nvs}
        \label{eq:total_loss}
    \end{align}
  }%
}
\noindent with the weights $\lambda_\text{segm}=10.0$, $\lambda_\text{smooth}=50.0$ and $\mu$ being view-dependent. 
$\mathcal{L}_\text{smooth}$ enforces smoothness in the rendered normal image. Both $\mathcal{L}_\text{RGB}$ and $\mathcal{L}_\text{segm}$ are supervised by $\{I^\text{obs}\}$ and $I^\text{ref}$, while $\mathcal{L}_\text{\nvs}$ is regularized by $\{\V{y}\}$.
See more details in \suppl.

\subsection{Hand-object alignment}
\label{sec:hand_ref}
Since the object lives in the \nvsm space and it does not have a metric scale, we need to estimate the scale of the object and align the hand and object based on reliable contact from image evidence.

\input{2_tex_FIG/reliable_contact}

\myparagraph{Visible contact}
Prior work~\cite{fan2024hold} identifies contact points as the closest hand--object vertex pairs.  
However, this strategy is unreliable when the object surface is heavily occluded or poorly reconstructed.  
Consequently, we enforce hand--object contact constraints only on \textit{visible contact pairs}.

Specifically, we take the vertices that frequently make contact on the fingertips (cyan mesh in \refFig{fig:reliable_contact}a)~\cite{hasson2019obman} as potential contacts, denoted by $\mathcal{V}_\text{h}^{\text{p}}$ (green dots in \refFig{fig:reliable_contact}a), while excluding those on the palm.
We project each $\mathbf{v}_h \in \mathcal{V}_\text{h}^{\text{p}}$ onto the image and label it as a \emph{visible contact candidate}, $\mathcal{V}_\text{h}^{\text{v}}$ (blue dots in \refFig{fig:reliable_contact}b), if its projection falls inside the hand mask; otherwise, it is marked \emph{object‑occluded} (red dots in \refFig{fig:reliable_contact}b) and discarded.
Some candidates in $\mathcal{V}_\text{h}^{\text{v}}$ are occluded by the front fingertips (yellow dots in \refFig{fig:reliable_contact}c) even though they lie inside the hand mask.  
Fingertips whose mask area exceeds a predefined threshold are deemed visible; we retain only the contacts on these fingertips, yielding the final reliable visible contacts $\mathcal{V}_\text{h}$.

For each contact in $\mathcal{V}_\text{h}$ we perform ray tracing to locate the corresponding object contact point $\mathcal{V}_\text{o}$, eliminating any non‑surface contacts.
This procedure yields a set of visible contact pairs $\{\mathcal{V}_\text{h}, \mathcal{V}_\text{o}\}$, which are then used to enforce reliable hand–object alignment.

\myparagraph{Alignment process}
For simplicity, we optimize only the hand while keeping the object fixed, using a contact-aware optimization procedure.
Our experiments show that allowing hand pose $\theta$ to optimize slightly increases the hand pose error (MPJPE: 7.83 mm), compared to the original estimates (MPJPE: 4.62 mm). 
Therefore, we optimize only the hand translation  $\V{t}_h\in \mathbb{R}^3$ and scale $s\in \mathbb{R}$ using the following objective function:
\begin{align}
\mathcal{L}_\text{h} &=
     \lambda_\text{contact}\mathcal{L}_\text{contact} + \lambda_\text{kpoints}\mathcal{L}_\text{kpoints} \notag \\ 
     &+ \lambda_\text{vsmooth}\mathcal{L}_\text{vsmooth} + \lambda_\text{penetr}\mathcal{L}_\text{penetr} 
\end{align}
where $\mathcal{L}_\text{contact}$ encourages a realistic hand-object contact relationship, $\mathcal{L}_\text{kpoints}$ enforces consistency in the hand joints \twoD projections, $\mathcal{L}_\text{vsmooth}$ ensures temporal smoothness of hand vertices, and $\mathcal{L}_\text{penetr}$ penalizes hand-object penetration. The corresponding weights are: $\lambda_\text{contact}=200.0$, $\lambda_\text{kpoints}=20.0$, $\lambda_\text{vsmooth}=20.0$, $\lambda_\text{penetr}=0.003$. See more details in \suppl.

%% file: 2_tex_FIG/method.tex
\begin{figure*}[t]
    \centerline{\includegraphics[width=\linewidth]{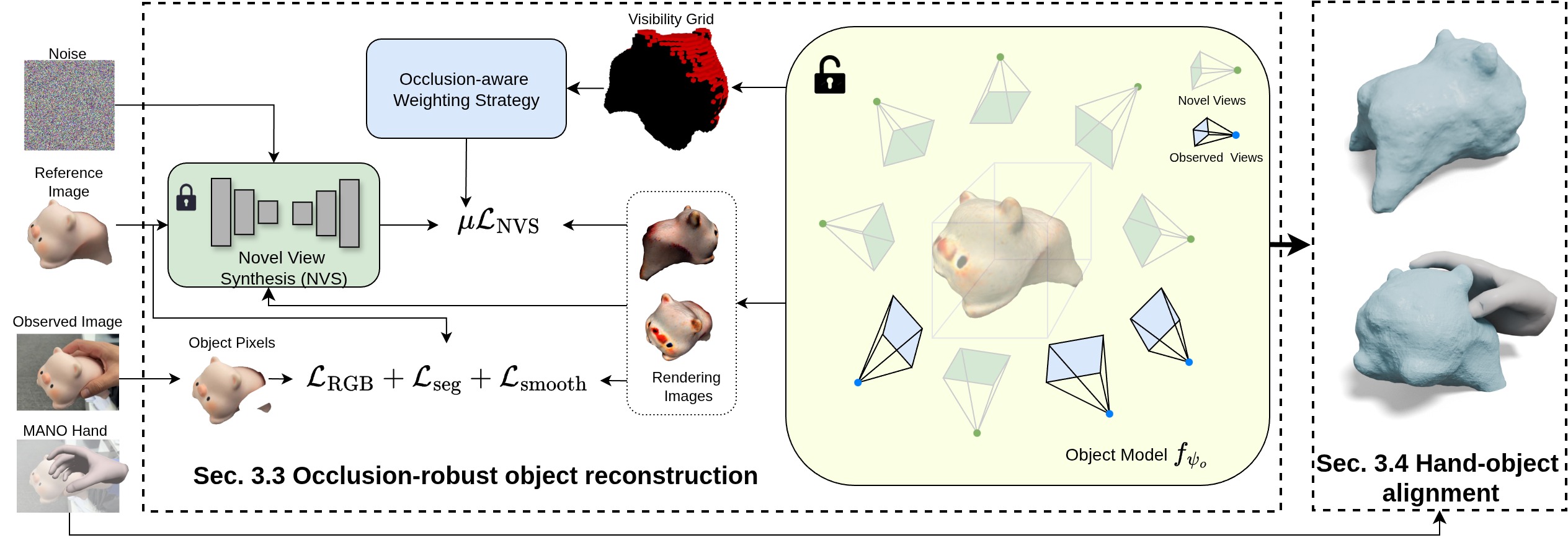}}
    \caption{
        \subtitle{Method overview}
        \methodname learns an implicit object field $f_{\psi_o}$ that reconstructs complete object geometry from a short video with partial visibility.
        We introduce a novel view synthesis (\nvs) loss $\mathcal{L}_\text{NVS}$ that leverages an \nvs diffusion prior to reconstruct unobserved object parts.
        Specifically, we render a novel view and take one denoising step of the \nvs model; $\mathcal{L}_\text{NVS}$ guides $f_{\psi_o}$ toward high-density regions to complete occluded areas.
        Updates from novel views are combined with RGB and segmentation supervision $\mathcal{L}_\text{RGB}$, $\mathcal{L}_\text{seg}$ and a smoothness regularizer $\mathcal{L}_\text{smooth}$ for both observed and reference images.
        A visibility-aware weighting factor $\mu$ emphasizes $\mathcal{L}_\text{NVS}$ for unobserved regions. 
        Finally, we align the hand to the completed object geometry to obtain an accurate hand–object spatial relationships.  
    }
    \label{fig:method}
\end{figure*}

%% file: 2_tex_FIG/reliable_contact.tex
\begin{figure}[t]
    \centerline{\includegraphics[width=\linewidth]{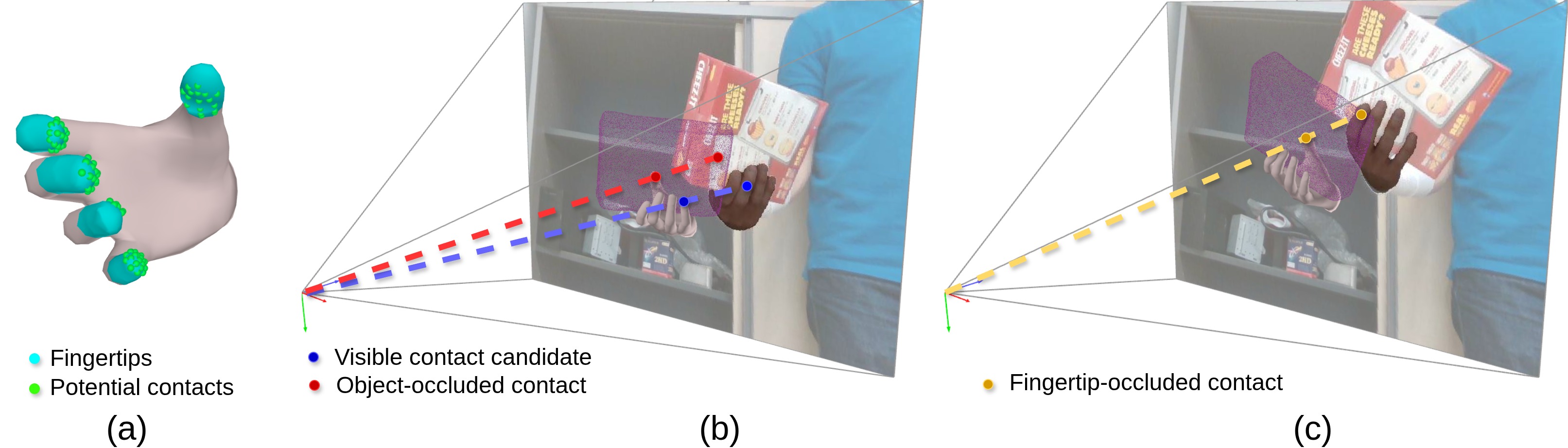}}
    \caption{
        \subtitle{Visible contact}  
        Our method first separate potential contacts on  fingertip into visible contact candidates and object-occluded contacts using hand mask information. 
        We then keep only those visible candidates whose projections are not occluded by other fingertips, yielding the final set of visible contact points.
    }
    \label{fig:reliable_contact}
\end{figure}%

%% file: 4_tex_Paper/4_experiments.tex
We evaluate our approach against state‑of‑the‑art baselines on the challenging task of \textit{partially visible hand–object reconstruction}.
The objective is to accurately recover both \threeD object geometry and hand pose from short monocular video sequences that provide only limited observation.

\subsection{Datasets}
\myparagraph{HO3D~\cite{hampali2023inhand}}
We evaluate our method using the HO3D-v3 dataset~\cite{hampali2020honnotate}, which contains RGB videos of hands interacting with YCB objects~\cite{ycb_object_2015}, along with annotations of hand poses and object poses.
We use the same sequences as HOLD~\cite{fan2024hold}, resulting in a total of 14 sequences.
Since our focus is on reconstruction from short videos, we divide long sequences into 30-frame clips (see SupMat), each providing limited observations.
The novel view synthesis (\nvs) reference frame is also used as input to iHOI~\cite{ye2022s} and EasyHOI~\cite{liu2024easyhoi}, which operate on a single image.

\myparagraph{In-the-wild sequences}
To evaluate real-world performance, we record short video clips of household objects in both indoor and outdoor environments, that do not reveal the complete objects.
We also include short videos from the internet to further assess our method's generalization.

\subsection{Metrics}
Following~\cite{fan2024hold, ye2023diffhoi, tatarchenko2019single}, we use the root-relative mean per-joint error (MPJPE) in millimeters to evaluate hand pose accuracy and the Chamfer distance (CD) in centimeters to assess object reconstruction quality. We also compute the F-score in percentage at 5mm (F5) and 10mm (F10) to compare object local shape details.
To measure the object's 3D pose and shape relative to the hand, we first subtract each object mesh using the predicted hand root position. We then compute the hand-relative Chamfer distance (CD$_h$) to measure the alignment accuracy.
To evaluate the size of the reconstructed object relative to its real size after hand alignment, we introduce the Relative Scale (RS) metric. In particular, $RS$ equals $\frac{1}{s} - 1~\text{if } s < 1$ and equals $1 - \frac{1}{s}, \text{if } s \geq 1$, where $s$ is the scale by performing ICP alignment  from the reconstructed object mesh to the \groundtruth. Intuitively, when $RS = 0$, the reconstructed object size matches the actual physical size. As $RS$ increases, the deviation from the real size becomes larger.

\input{2_tex_FIG/sota_compare}

\input{3_tex_TAB/object_recon}
\subsection{State-of-the-art comparison}

\refTab{tab:object_recon} presents a comparison between our method and state-of-the-art (SOTA) hand-object  reconstruction methods, including HOLD~\cite{fan2024hold}, EasyHOI~\cite{liu2024easyhoi}, DiffHOI~\cite{ye2023diffhoi}, and iHOI~\cite{ye2022hand}. 
All methods are template-free.
In particular, HOLD is a geometry-driven method that optimizes from long video inputs without using priors, while iHOI, EasyHOI, and DiffHOI are prior-driven methods. 
Our results indicate that \methodname achieves superior object reconstruction accuracy, as evidenced by the CD, F5, and F10 metrics. Among all baselines, HOLD overall struggles the most in this short video clip setting. 
Our method  has the best object scale estimate indicated by the RS metric.

\refFig{fig:sota_compare} shows a qualitative comparison of all methods to provide an intuition of the reconstruction results. 
From these results, we observe that prior-driven methods, \eg EasyHOI, DiffHOI, and iHOI, are capable of reconstructing complete object shapes even in regions that are occluded by the hand or not directly observed. 
This is in contrast to the non‑prior method, such as HOLD.
However, their reconstructed areas on the non-observed regions often exhibit distortions, particularly apparent in category-specific priors like those in iHOI and DiffHOI. 
EasyHOI generally does not reconstruct object surfaces that resemble those in the input images.
In contrast, \methodname combines the observations from the \rgb input with a category-agnostic \nvs prior, allowing for the most accurate and complete object reconstruction from short videos.
For short video sequences, artifacts in the reconstructed occluded regions can lead to significant pose alignment errors, as evidenced by the $CD_h$ values for the HOLD method in \refTab{tab:object_recon}. In contrast, our method reduces reconstruction noise in the non-observed object region, resulting in more stable and accurate hand alignment. Additional details are provided in \suppl.
In addition, 
\refFig{fig:sota_compar_frames} shows the video results across different frames, which indicate our method can generalize to bimanual setting in the rigid object sequences on ARCTIC~\cite{fan2023arctic} where there are more challenging hand-object occlusion, depth ambiguities, and more diverse hand pose variations.

\input{2_tex_FIG/sota_compare_rebuttal}

\subsection{Ablation}
To analyze the complementary roles of our novel view synthesis (\nvs) backbone, and \rgb image observations in the reconstruction quality in our short video clip setting, we conduct an ablation study in  \refTab{tab:ablation} based on different baselines set by changing the loss function in ~\refEq{eq:total_loss}:

\begin{itemize}  
    \item \textit{RGB} (row 2): Object reconstruction using the loss terms $\mathcal{L}_\text{RGB}, \mathcal{L}_\text{segm}, \mathcal{L}_\text{smooth}$. Both  $\mathcal{L}_\text{RGB}$ and $\mathcal{L}_\text{segm}$ are supervised only by the observed images $\{I^\text{obs}\}$.
    \item \textit{\nvs} (row 3): Reconstruction guided solely by \nvs-derived SDS loss condition on the reference image. In other words, we use all loss terms, but supervise $\mathcal{L}_\text{RGB}$ and $\mathcal{L}_\text{segm}$ only with the reference image $I^\text{ref}$ (no observed view supervision) and fix $\mu=1$.
    \item \textit{RGB+\nvs} (row 4): Integration of RGB inputs with \nvs via alignment without weighting strategy by setting $\mu$ to one. In other words, we use all loss terms, but fix $\mu=1$.
    \item \textit{Ours} (row 5): Object optimization with all loss terms. 
\end{itemize}

\myparagraph{Mutual benefits of \nvs model and \rgb observations}  
As shown in \refTab{tab:ablation}, combining \rgb image observations from the input video and the \nvs prior significantly improves reconstruction quality (see \textit{RGB} or \textit{\nvs} \vs \textit{RGB+\nvs}), evidenced by lower CD, hand-object alignment score CD$_h$, RS, and higher F5/F10 scores.
\refFig{fig:ablation} provides qualitative results to highlight the improvements. 
We see that only using \rgb, the model fails to  reconstruct regions occluded by hands or the object itself.  
Furthermore, only using \nvs model without incorporating other image evidence results in distorted and unrealistic object shapes (\eg, overly thick box/driller base, and unrealistic bottle geometry near occlusions).  
By combining with image observations and the \nvs prior, our model reconstructs more complete and accurate object shapes with minimal artifacts, closely matching the \groundtruth.

\input{3_tex_TAB/ablate}
\input{2_tex_FIG/ablate}
\myparagraph{Visibility-aware weighting strategy}
To assess the effect of our visibility-aware weighting strategy, we compare our full model \textit{Ours}  to the baseline \textit{RGB+\nvs} that omits this strategy while retaining both \rgb observations and \nvs priors (\refTab{tab:ablation}).
Quantitatively, the weighting strategy leads to improved reconstruction accuracy, as indicated by lower $CD$ and higher F5/F10 scores. Qualitatively, \refFig{fig:ablation} illustrates that removing this strategy results in surface distortions across both visible and occluded regions, and produces less smooth surfaces, especially in occluded areas.

%% file: 2_tex_FIG/sota_compare.tex
\begin{figure*}[t]
    \centerline{\includegraphics[width=0.90\linewidth]{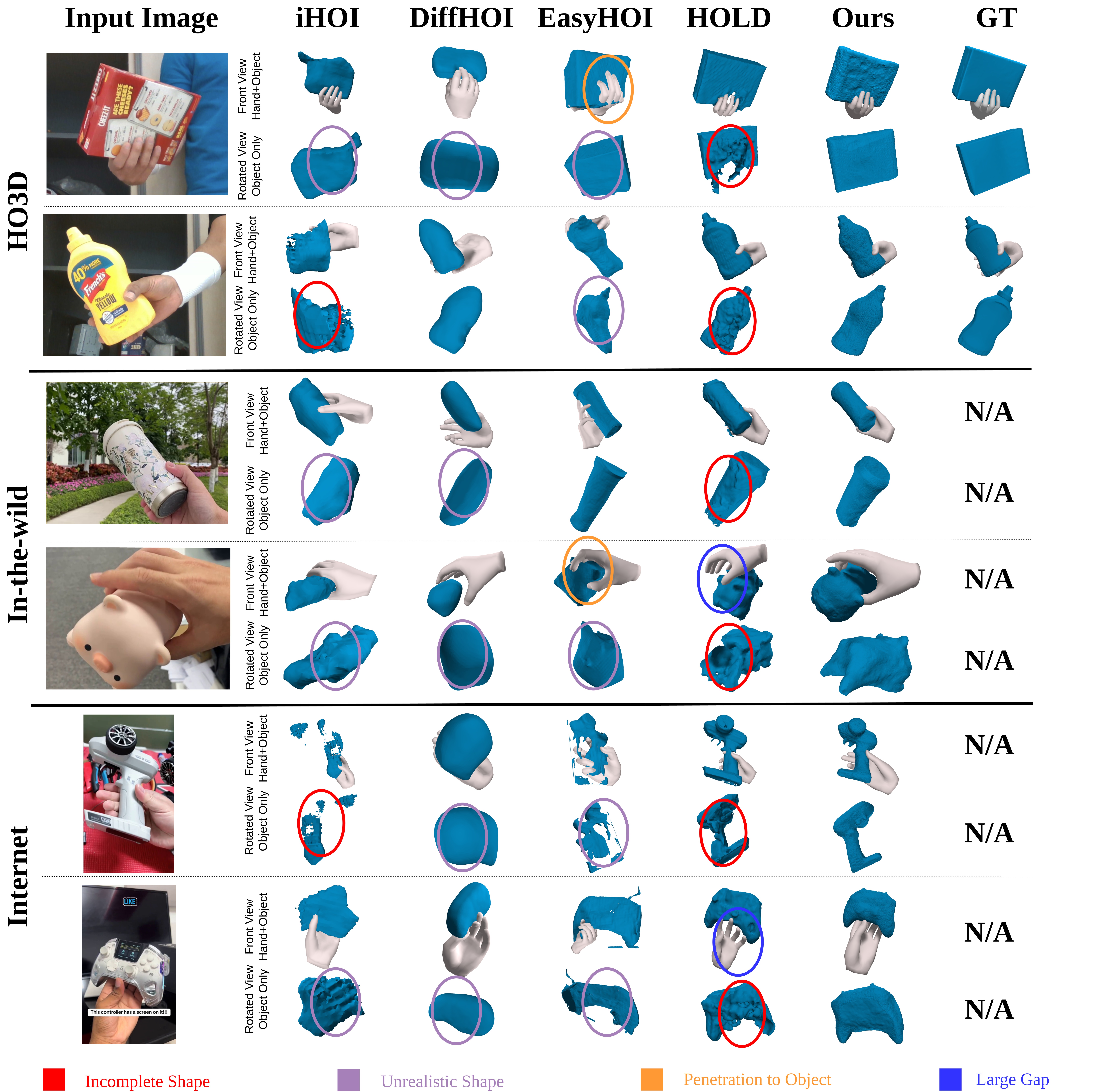}}%
    \caption{
        \subtitle{Qualitative comparison with \sota} 
        Reconstruction results from HO3D, in-the-wild, and internet short video sequences, comparing our method with \sota baselines in both hand-object front view and object only rotated view. Our approach excels in completing unobserved object regions and reconstructing complex shapes, even under severe hand-induced and object-self occlusion, yielding detailed geometry and realistic hand-object spatial relationships. 
        From the HOLD method column, it is evident that without an \nvs prior, it fails to reconstruct the complete object shape in unobserved regions and there is a large gap between hand and object. Even with an \nvs prior, the category-specific iHOI method struggles to recover full object geometry. EasyHOI, despite using an advanced \nvs model, often hallucinates unrealistic shapes and results in hand-object penetrations. DiffHOI, which integrates both prior-driven and geometry-driven approaches, tends to reconstruct only simple shapes.
    }
    \label{fig:sota_compare}
\end{figure*}%

%% file: 3_tex_TAB/object_recon.tex
\begin{table}[t]
\resizebox{1.0\linewidth}{!}{
\begin{tabular}{ccccccc}
\toprule
     & CD [cm] $\downarrow$ & F5 [$\%$]  $\uparrow$ & F10 [$\%$]  $\uparrow$  & MPJPE [mm] $\downarrow$ & CD$_h$ [cm] $\downarrow$ & RS $\downarrow$
\\
\midrule
iHOI~\cite{ye2022hand}                          & 2.37           & 35.78           & 62.11           & 27.75          &25.45           & 0.17         \\
DiiffHOI~\cite{ye2023diffhoi}                   & 2.30           & 39.59           & 64.49           & 16.02          &33.33           & 0.13        \\
    EasyHOI~\cite{liu2024easyhoi}               & 1.86           & 46.10           & 70.92           & 16.69          &19.55           & 0.28       \\
HOLD~\cite{fan2024hold}                         & 1.31           & 57.20           & 80.23           & 30.79          &21.28           & 0.62       \\
Ours                                            & \textbf{0.87}  & \textbf{69.72}  & \textbf{92.15}  & \textbf{4.62}  & \textbf{2.39}  & \textbf{0.11}   
\\ \bottomrule
\end{tabular}
}

\caption{
\subtitle{\sota comparison in hand-object reconstruction} We evaluate our method against four baselines on the HO3D dataset.}

\label{tab:object_recon}
\end{table}

%% file: 2_tex_FIG/sota_compare_rebuttal.tex
\begin{figure}[t]
    \centerline{\includegraphics[width=0.95\linewidth]{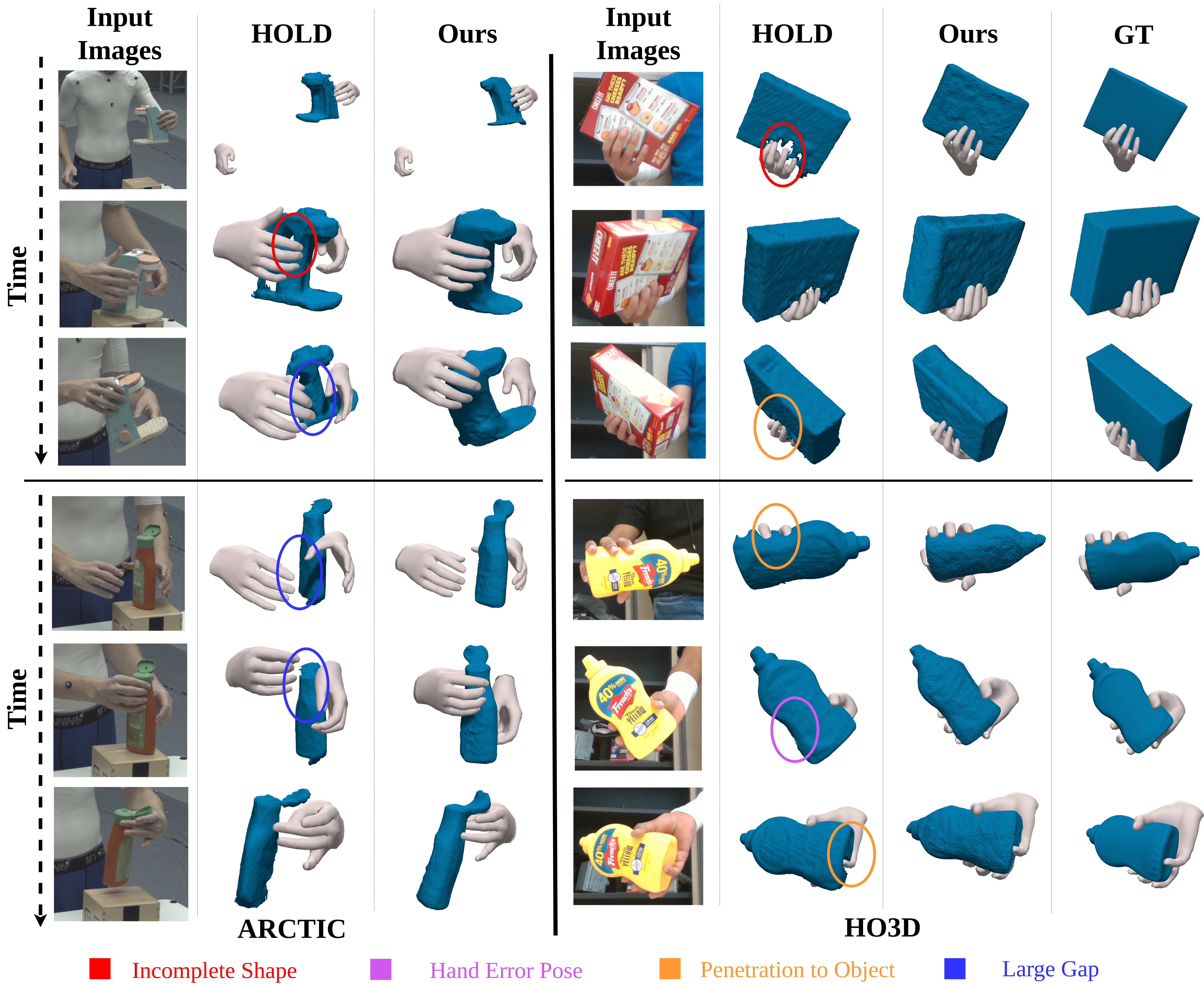}}%
    \caption{
        \subtitle{Video qualitative comparison with \sota } 
    }
    \label{fig:sota_compar_frames}
\end{figure}%

%% file: 3_tex_TAB/ablate.tex
\begin{table}
\resizebox{1.00\linewidth}{!}{
\begin{tabular}{cccc|ccccccc}
\toprule
& RGB & NVS & WS & CD [cm] $\downarrow$ & F5 [$\%$]  $\uparrow$ & F10 [$\%$]  $\uparrow$  & CD$_h$ [cm] $\downarrow$ & RS $\downarrow$
\\
\midrule
&{\checkmark} &{$\times$} &{$\times$}          & 1.37           & 55.02           & 78.24           & 3.21      & 0.20      \\
&{$\times$} &{\checkmark} &{$\times$}          & 1.54           & 55.59           & 78.97           & 11.97               & 0.26 \\
&{\checkmark} &{\checkmark} &{$\times$}        & 0.91          & 68.62           & 90.81           & 3.77               & 0.17  \\
&{\checkmark} &{\checkmark} &{\checkmark}     & \textbf{0.89}  & \textbf{69.72}  & \textbf{92.15}  & \textbf{2.39}     & \textbf{0.11}  \\
\bottomrule
\end{tabular}
}

\caption{
\subtitle{Ablation study} 
Integrating all observed RGB images into the NVS with the occlusion-aware weighting strategy (WS) significantly improves object reconstruction, as indicated by CD, F5, and F10. The refined object shape further enhances the hand-object relationships, as reflected in CD$_h$ and RS.
}
\label{tab:ablation}
\end{table}

%% file: 2_tex_FIG/ablate.tex
\begin{figure}[t]
    \centerline{\includegraphics[width=0.90\linewidth]{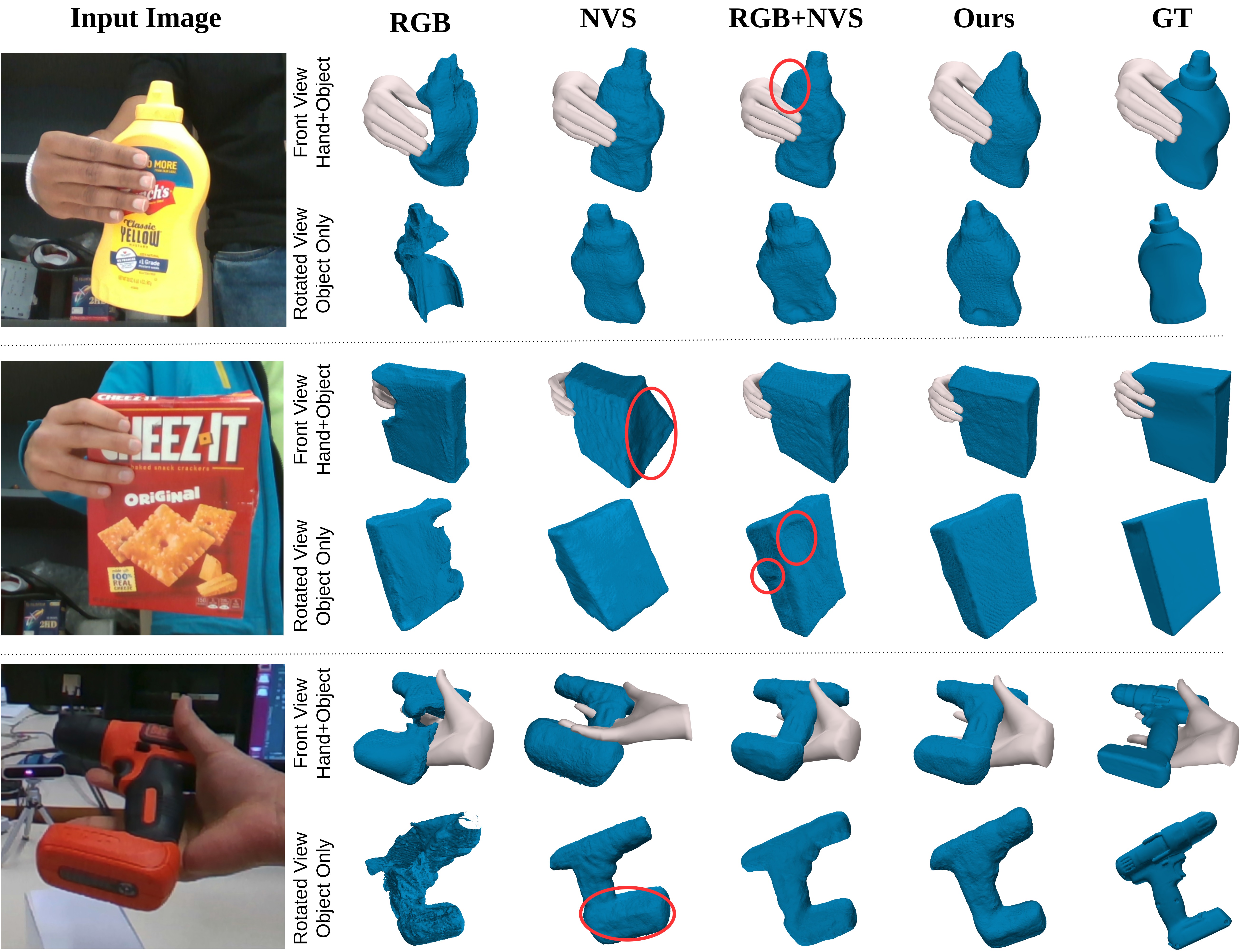}}
    \caption{
        \subtitle{Ablation Study}  Comparing \textit{RGB+NVS} and \textit{Ours} columns, the yellow bottle (row 1) exhibits significant distortion near the hand-occluded region, while the red box (row 2) shows numerous artifacts on observed surfaces in the \textit{RGB+NVS} column due to lack of a weighting strategy.
        Comparing \textit{NVS} and \textit{Ours} columns, the red box reconstructs an unrealistic shape on unobserved surfaces, and the driller (row 3) has an excessively thick base in the \textit{NVS} column.
        Comparing \textit{RGB} and \textit{Ours} columns, the object fails to reconstruct a complete shape in occluded regions.
    }
    \label{fig:ablation}
\end{figure}%

%% file: 4_tex_Paper/5_conclusion.tex
In this paper, we present \methodname, a hand-object reconstruction framework that incorporates novel view synthesis (\nvs) priors for reconstructing from short video sequences with significant occlusions.
By integrating \nvs into the reconstruction pipeline, \methodname enables accurate joint reconstruction of hands and objects even in such challenging interaction scenarios.
Qualitative and quantitative results show that \methodname outperforms \sota methods without requiring full object or hand visibility in both controlled and in-the-wild settings.\\

\myparagraph{Acknowledgment} We thank Muhammed Kocabas, Xu Chen, Bonan Liu for detailed discussions and insightful feedback, Handi Yin for support and International Max Planck Research School for Intelligent Systems (IMPRS-IS) for supporting Maria Parelli.